\documentclass[journal]{IEEEtran}
\usepackage{amsmath,amsfonts}
\usepackage{algorithmic}
\usepackage{algorithm}
\usepackage{array}
\usepackage[caption=false,font=normalsize,labelfont=sf,textfont=sf]{subfig}
\usepackage{textcomp}
\usepackage{stfloats}
\usepackage{url}
\usepackage{verbatim}
\usepackage{graphicx}
\usepackage{cite}
\usepackage{multirow}

\usepackage{pifont}
\newcommand{\cmark}{\ding{51}}%
\newcommand{\xmark}{\ding{55}}%

\usepackage{caption}
\begin{document}

\title{Confidence-based Intent Prediction for Teleoperation in Bimanual Robotic Suturing}


\author{Zhaoyang Jacopo Hu, Haozheng Xu, Sion Kim, Yanan Li, Ferdinando Rodriguez y Baena,
and Etienne Burdet

\thanks{This work was supported by the EPSRC and Intuitive Surgical in the form of an industrial CASE studentship.
\it{(Corresponding authors: Zhaoyang Jacopo Hu, Ferdinando Rodriguez y Baena, Etienne Burdet).}}
\thanks{Zhaoyang Jacopo Hu, Sion Kim and Ferdinando Rodriguez y Baena are with the Department of Mechanical Engineering, Imperial College London, SW7 2AZ, UK (e-mail: jacopo.hu20@imperial.ac.uk, f.rodriguez@imperial.ac.uk). Haozheng Xu is with the Department of Surgery \& Cancer, Imperial College London, SW7 2AZ, UK. 
Yanan Li is with the Department of Engineering and Design, University
of Sussex, Brighton, BN1 9RH, UK.
Etienne Burdet is with the Department of Bioengineering, Imperial College London, W12 0BZ, UK. (e-mail: e.burdet@imperial.ac.uk).}
}



\maketitle

\begin{abstract}
Robotic-assisted procedures offer enhanced precision, but while fully autonomous systems are limited in task knowledge, difficulties in modeling unstructured environments, and generalisation abilities, fully manual teleoperated systems also face challenges such as delay, stability, and reduced sensory information.
To address these, we developed an interactive control strategy that assists the human operator by predicting their motion plan at both high and low levels. At the high level, a surgeme recognition system is employed through a Transformer-based real-time gesture classification model to dynamically adapt to the operator's actions, while at the low level, a Confidence-based Intention Assimilation Controller adjusts robot actions based on user intent and shared control paradigms. 
The system is built around a robotic suturing task, supported by sensors that capture the kinematics of the robot and task dynamics. 
Experiments across users with varying skill levels demonstrated the effectiveness of the proposed approach, showing statistically significant improvements in task completion time and user satisfaction compared to traditional teleoperation.

\end{abstract}


\section{Introduction}
\IEEEPARstart{I}{n} traditional teleoperation the human operator fully controls the robot's movements \cite{sheridan1992telerobotics}.
While this offers clear roles and responsibilities, it under-utilizes both the human and robot capabilities. Robots like the da Vinci Surgical System are equipped with sensors and models offering valuable local information inaccessible to the human operator, such as during visual occlusions or operations with different sensory modalities.  
By spanning across the spectrum between traditional fully manual teleoperation and full autonomy, shared control leverages the benefits of both to enhance teleoperation with the robot's sensory data and control \cite{jarrasse2014slaves}. 
While demonstrated for suturing assistance \cite{selvaggio2019haptic, fontanelli2018comparison}, these methods overlook the impact on positional uncertainty, environmental unknowns, or instrument errors.
For example, robotic surgery cameras are frequently occluded by body tissues or parts of the robot \cite{tukra2021see}.

\begin{figure}[t] 
\centering 
\includegraphics[height=10.1cm, trim={0.95cm 0.0cm 0.95cm 0cm },clip]
{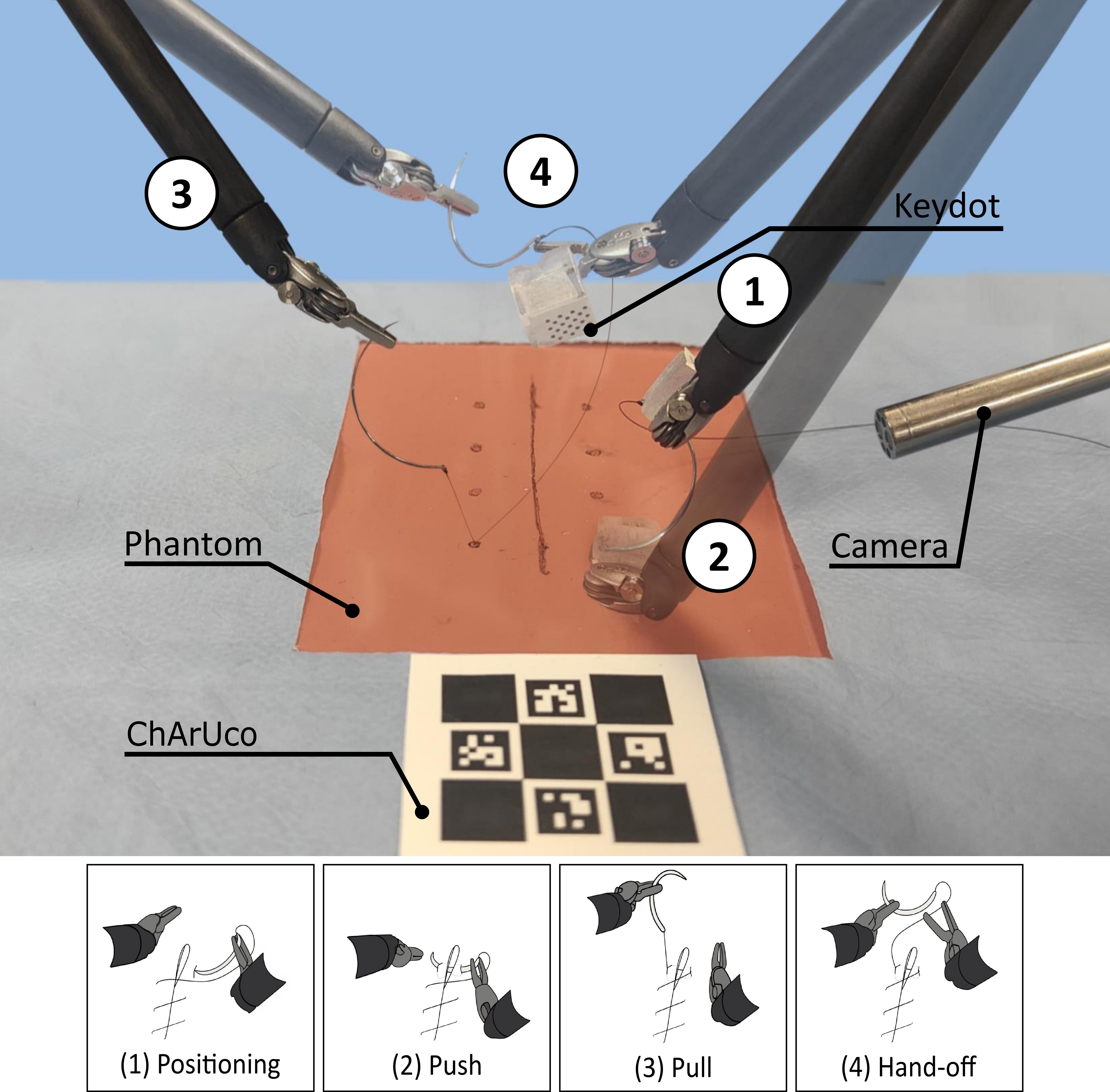}
\caption{Bimanual robotic suturing with four key surgemes: (1) Needle tip positioned at the entry point; (2) Needle pushed/inserted reaching the exit point; (3) Needle pulled out; (4) Needle passed to the starting manipulator. 
} 
\label{fig: cover_picture}
\end{figure}

\IEEEpubidadjcol 

To address these shortcomings, this paper develops an interaction control framework for the da Vinci Surgical System, framing shared control as an information exchange between human user and robot agent based on the user's intent \cite{li2022review}.
Inspired by studies on human-human collaboration \cite{ganesh2014two, takagi2019individuals}, this approach shares not only current positions but also motion plans between the partners, enabling them to anticipate each other's actions and optimize movement control \cite{takagi2017physically}.
At \textit{task level}, the robot identifies the operator's desired high-level actions through specific hand gestures, or \textit{surgemes} \cite{van2021gesture}. These surgemes can be used to infer the surgeon's task throughout the procedure (Fig.\,\ref{fig: cover_picture}). For instance, online gesture classification has been used to automate camera motion \cite{pasini2023grace} or switch assistance modes \cite{rakita2019shared}.
After recognizing a surgeme, the robot estimates the operator's motion intention at a lower \textit{control level} to adapt seamlessly during collaborative tasks, akin to the interaction observed in connected humans carrying out a tracking task \cite{takagi2017physically, li2022review}. In \cite{takagi2020flexible, chen2024human, zheng2024user, hu2023towards} the authors suggest that adaptation and incorporation of motion intentions can minimize the operator's effort.
The \textit{intention assimilation controller} (IAC) \cite{takagi2020flexible} enables the robot to identify the operator's motion plan and combine it with its own based on task requirements and sensor input. 
However, \cite{takagi2020flexible} did not specify the criteria for transitioning between interaction strategies, which may be regulated based on the robot's confidence in its sensors \cite{li2022review, kam2021confidence}. Confidence scores can help avoid collisions \cite{chernova2009interactive} or improve trajectory tracking \cite{saeidi2018confidence}, facilitating human-robot negotiation based on information received. In this paper, we assume camera data to be more accurate than forward kinematics for position estimation, but prone to occlusions and fluctuations.

We introduce an interaction control framework that integrates task-level prediction with \textit{confidence-based IAC} (C-IAC), implemented on the da Vinci Research Kit (dVRK). The dVRK offers high-definition cameras but limited sensory feedback, making tasks like suturing tedious, time consuming, and attention-intensive \cite{ostrander2024current}. We hypothesize that, by combining sensing and motion planning of the human and robot, one can use sensory augmentation to enhance suturing performance and comfort.
To test this, we developed a kinematic suturing dataset and trained a Transformer model to infer suturing activity in real-time on the dVRK by decomposing it into surgemes. We also developed a mechanical device to automatically orient the needle relative to the manipulator. We then designed an IAC-based controller with confidence level computed from Bayesian inference analyzing kinematic and camera data. Two user studies evaluated the entire system's performance.
The main contributions of this paper are as follows:
\begin{itemize}
  \item A novel needle holder for facilitated grasping.
  \item Integration of shared control with a gesture recognition model for bimanual robotic suturing using the dVRK.
  \item Implementation of a dynamic parameter to compute the confidence component based on Bayesian inference.
  \item Development of a confidence-based IAC controller used to teleoperate the da Vinci robot under shared control.
  \item User studies to evaluate the performance of the \text{C-IAC}.
\end{itemize}



\begin{figure*}[t] 
\centering 
\includegraphics[height=3.3cm, trim={1.6cm 4.3cm 1.6cm 4.4cm},clip]{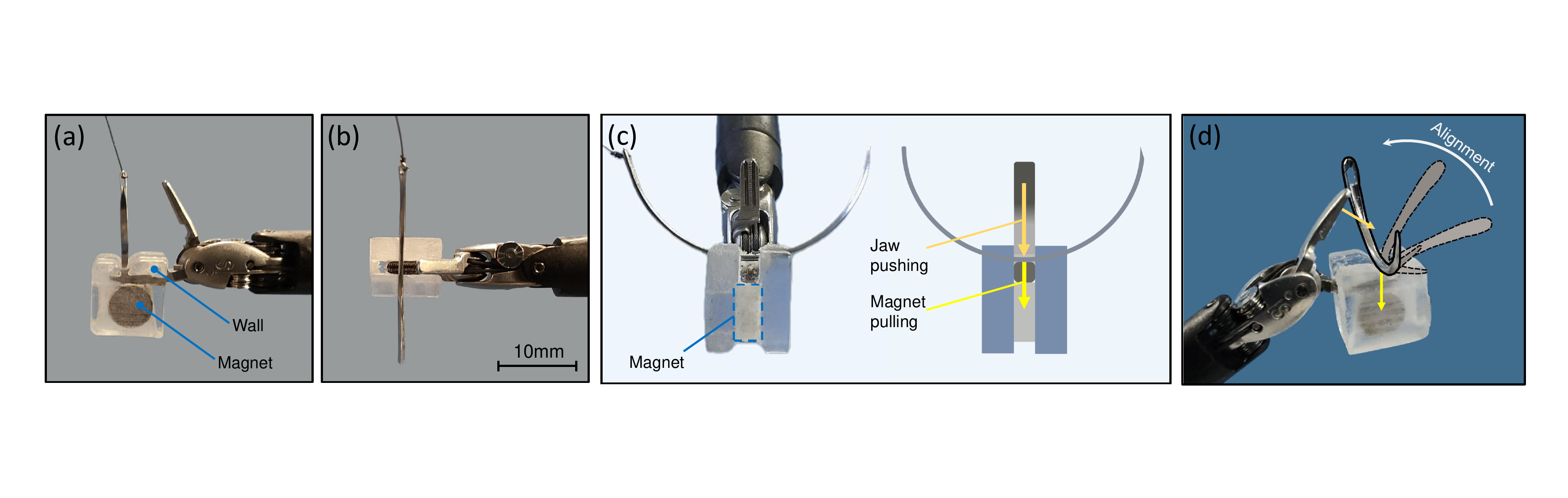}
\caption{Design of the  Compact Holder for Enhanced Needle Alignment (CHENA). (a-b) The structure comprises of walls that create a concavity where the needle is attracted and maintained fixed by a magnet. (c-d) The needle self-aligns using to two independent acting forces: the jaw pushing and the magnet pulling the needle towards the cavity. } 
\label{fig: CHENA design}
\end{figure*}

\section{Surgical Setting} \label{Surgical setting}

\subsection{Compact Holder for Enhanced Needle Alignment}
During suturing, operators often require multiple manipulations of the needle to achieve the optimal orientation \cite{guni2018development, pedram2020autonomous}. 
Previous attempts in solving this problem include  \cite{sen2016automating}, which created a component whose geometry enabled needle alignment and that could be easily attached to the existing surgical tool without requiring a complete replacement. However, the device increases the size of the end-effector, making needle handling awkward.
We developed a low-cost \textit{Compact Holder for Enhanced Needle Alignment} (CHENA), improving prior designs \cite{sen2016automating, pedram2020autonomous} by facilitating needle grabbing and holding (Fig.\,\ref{fig: CHENA design}). The aligner in \cite{sen2016automating} was intended for autonomous suturing and required a large rear wall and catching area to guide the needle. The necessity of these parts force the components to extrude over the opposite side of the surgical tool, limiting its maneuverability and maximum jaw open angle.
In contrast, CHENA self-aligns and positions the needle without the need for large extrusion in the catching area by using a minimalistic geometry and a magnet to create a driving force for the needle to align even without closing the jaws. 



\subsection{dVRK Setup}

For the experiments, we use the dVRK equipped with two \textit{Patient Side Manipulators} (PSMs): right PSM1 with DeBakey Forceps and CHENA, and left PSM2 with Large Needle Drivers.
The needle used is a $1/2$ circle, 24 mm chord length with triangular cutting point.
Robot control is performed in Cartesian space via the dVRK software.
Two sigma.7 hand interfaces control the robot using the TCP/IP protocol. 
As in \cite{saracino2019haptic, hubens2003performance}, we renounce to binocular vision and provide a 2D high resolution display to the human user using a RealSense camera, with a 50\,ms delay introduced by the teleoperation to observe the effects on performance between controllers \cite{richter2019motion}.
Latency contributes to the complexity and unpredictability of the task and surgeon's performance and is considered one of the biggest hurdles to enable telepresence \cite{sung2001robotic}. \cite{richter2019motion, xu2014determination} showed that a latency below 200\,ms is ideal for telesurgery, but even 50\,ms can impact performance.
This setup aims to emphasise the disparity in sensory information between human and robot, impacting performance outcomes. 
We maintain pedals for clutching and control of the end-effector orientation.

A surgical phantom is created to perform the suturing task using a 1\,mm thick Ecoflex 00-50 layer, replicating the setup in \cite{gao2014jhu}. Similarly, the wound and entry points are manually marked, measured and recorded relative to a ChArUco marker on the phantom.  A keydot marker on the CHENA tracks the PSM1 and needle pose. 
Marker tracking relies on the da Vinci Si endoscope, which provides high-resolution images.

\section{Kinematic Gesture Recognition}

The Transformer \cite{vaswani2017attention} improves upon existing gesture recognition models with its ability to process and analyze kinematic data in parallel. 
Encoder-only configurations of the Transformer have proven effective in classification tasks \cite{lin2022survey, zabihi2023trahgr}, suggesting their potential applicability in surgical gesture classification. Therefore, we implement a real-time gesture recognition system that leverages the encoder-only Transformer model. This system classifies gestures by processing a continuous window of kinematic data from the two PSMs and two sigma.7 hand interfaces.

\subsection{Suturing Dataset} 

A major obstacle in robotic surgery gesture recognition is the scarcity of annotated datasets \cite{van2021gesture}, with one of the few being JIGSAWS \cite{gao2014jhu}, which is widely used as a benchmark dataset. To accommodate our specific settings, which are slightly different from JIGSAWS, we constructed our own
kinematic dataset made of bimanual suturing motions, referred to as \textit{Suturing Task in Imperial's Tracking Collection for Hamlyn's Evaluation Set} (STITCHES). 
It includes kinematic and video data recorded at 20\,Hz from a user with over 900 hours of dVRK experience. 
The dataset captures movements from the two PSMs and two sigma.7s, totaling 19 kinematic features per device: end effector position (3), rotation matrix (9), linear velocity (3), angular velocity (3), and gripper angle (1). Gesture labels, manually annotated using video references, add a final feature, resulting in 77 features. We collected 10 recordings, each having four sutures, or throws, defined as the combination of surgemes completing one loop.

\subsection{Gesture Recognition Algorithm} \label{Sec.: Transformer}

The kinematic window captures linear velocity (3), rotational velocity (3), and gripper angle (1) from each of the four devices at 20\,Hz, resulting in 28 features per time step. This window spans 60 time steps (3 seconds) and is fed into a classification network composed of two Transformer encoder blocks, each with 
attention heads,
followed by two fully connected layers of 64 units each and a softmax activation. 
To improve stability in real-time, we use an exponential moving average (EMA) window of size 10 over the output probabilities and a 0.8 probability threshold on averaged probabilities. 
The EMA filters noise and transient errors, ensuring stable gesture predictions while adapting quickly to new gestures by weighting recent data more heavily. The threshold ensures that only high-probability gestures are recognized. 
We validated the model using 5-fold cross-validation, dividing the 10-recordings dataset into five folds, each containing two recordings. 

\subsection{Suturing Gesture Classification}

Suturing is a fundamental laparoscopic surgery task \cite{ritter2007design}. \cite{gao2014jhu} identified 15 different gestures $G$ in robotic surgery, 10 of which apply to suturing. However,  \cite{van2021gesture, sen2016automating} suggested that suturing could be represented with just 4 more general fundamental gestures. 
We demonstrate the viability of this approach by using the JIGSAWS dataset \cite{gao2014jhu}, which has been widely recognized as a benchmark for the dVRK, and proposing two strategies to group the original gesture labels $G$ from JIGSAWS into 5 classes (Table \ref{demo-table1}). 
Both include the fundamental gestures (1) \textit{Positioning}, (2) \textit{Push}, (3) \textit{Pull}, and (4) \textit{Hand-off}, as illustrated in Fig.\,\ref{fig: cover_picture}, and an additional \textit{Other} class to group the rest of the gestures. In strategy 1, only the original four gestures $G$ are labeled with the corresponding surgeme while the rest are labeled \textit{Other}.
Strategy 2 groups gestures similar in motion to the four original gestures and the rest is labeled \textit{Other}.
If the Transformer-based model can classify both strategies with similar accuracy, then this enables us to verify whether four surgemes can represent suturing adequately.
We can then implement strategy 2, which more broadly represents each gesture, in the shared control system's Transformer model.
\begin{table}[!h]
\caption{\label{demo-table1} Proposed Surgeme Labels (\textnormal{a-d}) using JIGSAWS}
\begin{center}
\begin{tabular}{|c|c|c|}
\hline \textbf{Surgeme} & \textbf{Strategy 1} & \textbf{Strategy 2} \\
\hline \begin{tabular}{c} 
(1) 
\end{tabular} & G2 & G2, G5 \\
\hline \begin{tabular}{c} 
(2) 
\end{tabular} & G3 & G3 \\
\hline (3)  & G6 & G6, G10 \\
\hline \begin{tabular}{c} 
(4) 
\end{tabular} & G4 & G4, G8 \\
\hline Other & G1, G5, G8, G9, G10, G11 & G1, G9, G11 \\
\hline
\end{tabular}
\end{center}
\end{table}

\section{Confidence-based IAC} \label{Sec: Confidence-based IAC}
\subsection{Intention Assimilation Controller}\label{Sec: Confidence-based IAC - A}

IAC enables the robot to extract the human's motion plan from the interaction force and then combine it with its own plan \cite{takagi2020flexible}. 
Given the operator's applied force $u_h$, system position $x$, and velocity $\dot{x}$, IAC predicts the human's motion plan and ensures stable human-robot interaction \cite{takagi2020flexible}:
\vspace{-2pt}
\begin{equation}\label{dynamics_model}
u_h=-L_1(x-\tau)-L_2 \dot{x}
\end{equation}
where $L_1$, $L_2$ are stiffness and viscosity matrices \cite{takagi2020flexible}. The robot target $\tau$ is extracted from Equation \ref{dynamics_model} during the movement using a Kalman filter to compute the predicted state estimate of the human target $\hat{\tau}_h$. 
The robot is then controlled via:
\vspace{-2pt}
\begin{equation}\label{shared_eq}
\begin{aligned}
& \tau=\lambda\tau_r + (1-\lambda)\hat{\tau}_h
\end{aligned}
\end{equation}
where the robot target is the convex combination of the estimated human target $\hat{\tau}_h$ and its own target $\tau_r$, selected based to its task and sensing information.
The confidence factor $\lambda$ balances the robot's reliance on information from its target $\tau_r$ versus the human target ${\tau}_h$, influencing assistance behavior during interaction.

\begin{figure}[h] 
\centering 
\includegraphics[height=4.3cm, trim={16cm 1cm 16cm 4.0cm },clip]{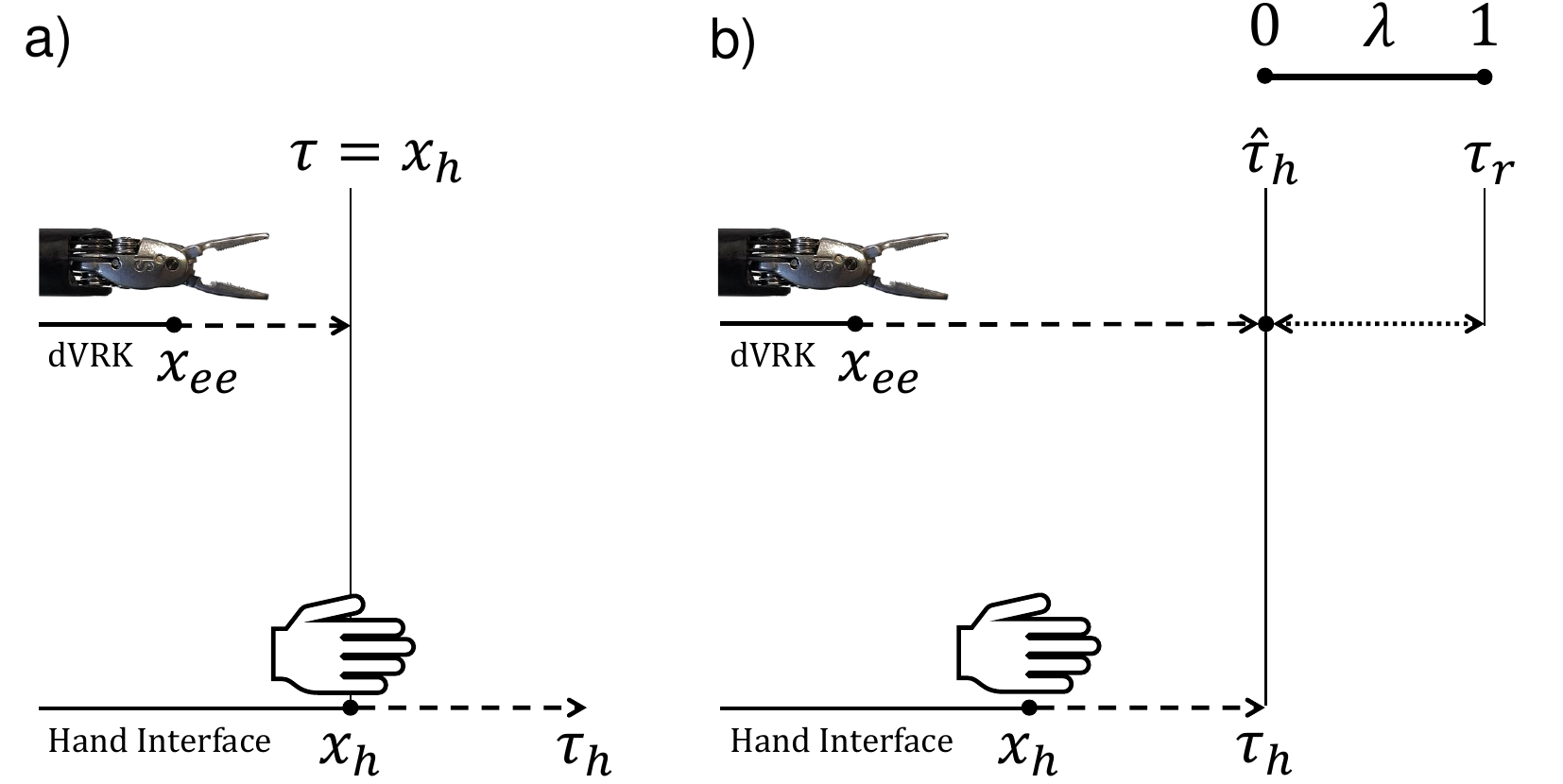}
\caption{Comparison between a) traditional and b) C-IAC teleoperation. While traditional teleoperation has inherent delay to reach the current position of the user's hand, C-IAC directly moves towards the predicted position. $\lambda$ is used in the shared control scheme to adjust the assistance. With high confidence ($\lambda\approx1$), delay is minimized as the robot to follow its own target.} 
\label{fig: trad_vs_IAC}
\end{figure}

When $\lambda=0$, the robot considers its own information unreliable and fully follows the predicted human target position $\hat{\tau}_h$.
$0<\lambda<1$ sets a cooperation between human and robot control inputs, considering both of their targets. This allows to benefit from the speed of reaching the predicted human target $\hat{\tau}_h$ instead of the current position $x_h$ while dynamically adjusting confidence based on the sensory information detected by the robot.
At $\lambda=1$, the robot disregards the human interaction, adopting a coactive behavior to achieve its own target, assuming sufficient information to complete tasks without teleoperation delays.
To support the user, $0\leq\lambda\leq1$ must be ensured. 
The implementation of C-IAC compared to traditional teleoperation is shown in Fig.\,\ref{fig: trad_vs_IAC}.
Dynamic and continuous adjustment of $\lambda$ is key for smooth and situation-specific human-robot interaction. This method allows to optimize the control authority in real-time according to the system's confidence on either the human or robot.


\subsection{Confidence with Bayesian Inference}\label{Bayesian Inference}

We use vision and proprioception as sensory inputs for the robot. Visual information comes from tracking markers placed on the robot's manipulator and phantom (keydot $kd$ and ChArUco $ch$, respectively). Proprioception is estimated using the robot joint angles and is consistently available, unlike vision that can be obstructed, leading to a loss of tracking.

We define confidence as the system's trust in the sensors' data.
As shown in \cite{guo2021modeling}, Beta distributions effectively infer the human's trust in the robot's performance by analyzing the robot's performance history.
Additionally, the distribution is bounded within the the interval $[0,1]$ which is consistent with the confidence boundaries to achieve the interaction behavior needed. Furthermore, the distribution dynamics enables the property of influencing the confidence at the current moment $i$ by previous time steps $i - 1$ \cite{lee1992trust}.

Therefore, we use Bayesian inference to calculate the trust in the visual tracking data continuously, prioritizing recent observations.
The Beta distribution, with its shape parameters $B(\alpha,\beta)$,
updates dynamically as new sensory data is received.
We use the Bayesian inference from \cite{guo2021modeling} to estimate the system's confidence in its vision.
Effectively, visual information yields a boolean message for both keydot and ChArUco markers, indicating whether the information is received or absent at each time step. This determines an arbitrarily assigned performance $p$: 0 if the system lacks confidence, and 1 if the system has confidence  (i.e. $kd$ $\vee$ $ch$, illustrated in Table \ref{table: message weight class}).
The performance history in a window of length $n$ is denoted as $P=\left\{p_1, \ldots, p_n\right\}$.
\vspace{-2pt}
\begin{table}[!h]
\centering
\caption{\label{table: message weight class} 
Visual Information Performance Assignment}
\begin{tabular}{|c|c|c|c|c}

\hline 
\textbf{kd} &
\textbf{ch} &
\textbf{Performance $p_i$} &
\textbf{Gain $w_{p_i}$} \\
\hline  \xmark & \xmark  & 0 & $w_0$ \\
\hline  \cmark & \xmark  & 1 &  \\
\cline{1-3} \xmark & \cmark & 1 & $w_1$ \\
\cline{1-3}  \cmark & \cmark & 1 &  \\

\hline
\end{tabular}
\end{table}



As the marker tracking performance is received, at each $i$th time step the system's trust dynamics is calculated as:
\begin{equation}
\begin{aligned}
B_i\left(\alpha_i, \beta_i\right) \quad & \quad \alpha_i = \begin{cases} 
\alpha_{i-1} + w_1, & \text{if } p_i = 1 \\
\alpha_{i-1}, & \text{if } p_i = 0
\end{cases} \\
& \quad \beta_i = \begin{cases} 
\beta_{i-1} + w_0, & \text{if } p_i = 0 \\
\beta_{i-1}, & \text{if } p_i = 1
\end{cases}
\end{aligned}
\end{equation}
where $\alpha_i$\ and $\beta_i$ are the shape parameters, and $w_{p_i}$ are the respective gains.
The posterior mean is then used to estimate the system's confidence:
\vspace{-2pt}
\begin{equation}\label{alpha_mean}
\lambda = E\left(B_i\right) = \frac{\alpha_i}{\alpha_i+\beta_i}
\end{equation}

Equation \ref{alpha_mean} ensures a continuous value within $[0,1]$ to assess system confidence in its sensory data. The value can be scaled to match all behaviors described in Section \ref{Sec: Confidence-based IAC - A}, depending on the task needs.
Notice that the confidence primarily dependents on vision rather than on proprioception. We assume that, when available, visual information of the end-effector position is more reliable than proprioception,  as the dVRK's cable-driven system suffers from large positioning errors due to cable elasticity and long kinematic chains \cite{haghighipanah2015improving, pedram2020autonomous}.
We empirically tuned the gains $w_{p_i}$ and assigned positive performance when $kd$ or $kc$ was detected, and zero otherwise (Table \ref{table: message weight class}). 



\subsection{Shared Control Paradigms}\label{shared control paradigms section}

\begin{figure}[h] 
\centering 
\includegraphics[height=2.6cm, trim={0cm 0cm 0cm 0.5cm },clip]{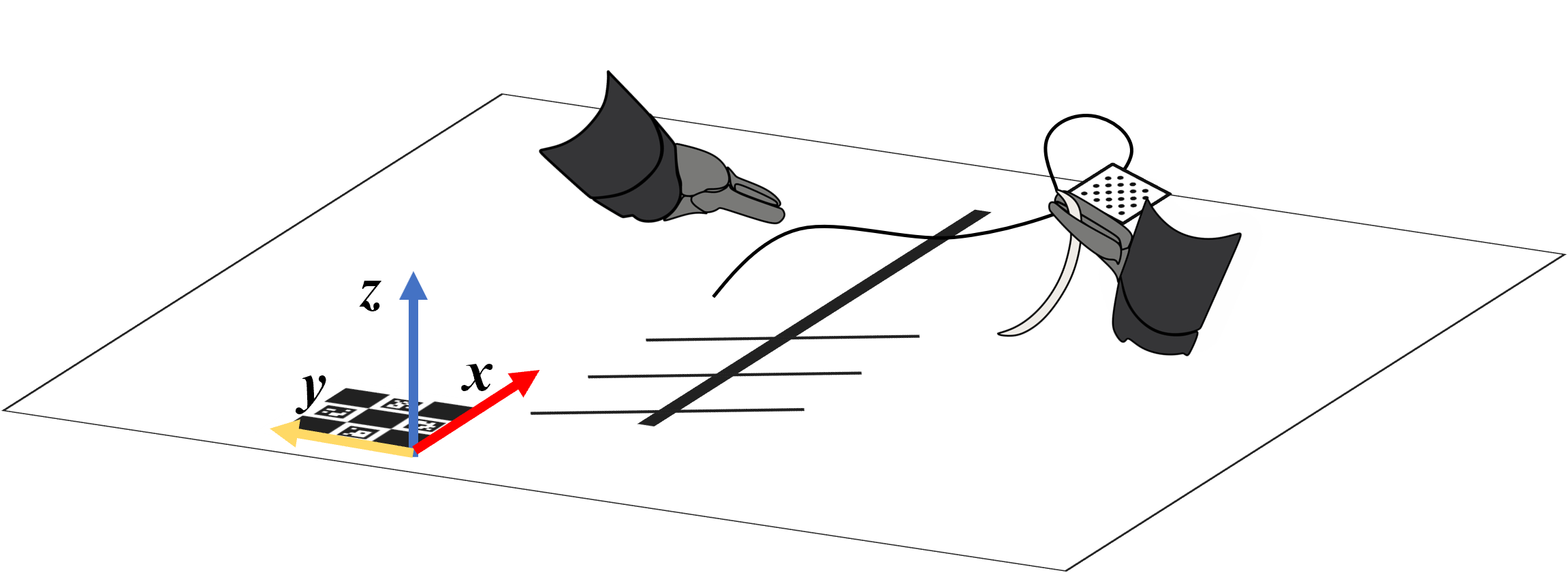}
\caption{Task frame position on the tissue surface using a ChArUco with x-axis parallel to the wound, y-axis perpendicular to it, and z-axis perpendicular to the tissue surface. During suturing, the robot refers to the task frame Cartesian coordinates to determine wound orientation and insertion point.} 
\label{fig: tissue_global_frame}
\end{figure}

We propose the following shared control paradigms, generalized into the four fundamental surgemes and a general one. The constraints and targets in these paradigms have been developed in consultation with a neurosurgeon from Imperial College London. The following terminology represents quantities measured relative to the task frame in Fig.\,\ref{fig: tissue_global_frame}, defined with respect to the manipulator tracked by the keydot:

\begin{itemize}
  \item $\tau =( x, y, z) \in \mathbb{R}^3$: target position computed by C-IAC.
  \item $\lambda =( x, y, z) \in \mathbb{R}^3$: confidence applied on end-effector.
\end{itemize}
\vspace{-15pt}

\subsubsection{Positioning}
The user aligns the needle's plane perpendicular to the wound, ensuring the tip touches the insertion point at a fixed height. 
Since the entry point can vary along the tissue near the wound, the manipulator moves freely along a plane parallel to the tissue and at fixed target height. 
\vspace{-2pt} 
\begin{gather}\label{eq. positioning}
\tau= ( \tau_h^x, \tau_h^y, \tau^z) \,,\quad
\lambda = ( 0, 0, \lambda^z)
\end{gather} 
As the needle plane needs to be perpendicular to the wound for an effective needle insertion, we allow the human to press a pedal to trigger autonomous orientation of the end-effector.

\subsubsection{Push}
The needle plane must remain perpendicular to the tissue surface and the wound direction. To prevent laceration, the manipulator movements along the wound axis are restricted, while having free motion in other directions. This enables to perform a correct circular motion for the needle to emerge from the opposite side of the wound.
\vspace{-2pt}
\begin{gather}\label{eq. push}
\tau = ( \tau^{x_j}, \tau_h^y, \tau_h^z ) \,,\quad
\lambda = ( \lambda^x, 0, 0 )
\end{gather} 
where $x_j$ is the $j$th entry point position in $x$ during suturing.
As in \textit{Positioning}, the end-effector orientation can be controlled using a pedal to autonomously achieve perpendicularity.

\subsubsection{Pull}
When one hand is stationary and the other moves with higher velocity, stability in the stationary hand improves task performance \cite{rakita2019shared}. In suturing, this corresponds to \textit{Pull}, where the right hand is kept steady above the tissue, allowing users to focus on pulling with the left. To facilitate the next throw, the robot target is set to the next entry point position.
\vspace{-2pt}
\begin{gather}\label{eq. pull}
\tau= ( \tau^{x_{j+1}},\tau^y, \tau^z ) \,,\quad
\lambda = ( \lambda^x, \lambda^y, \lambda^z )
\end{gather}

\subsubsection{Hand-off} The hand-off of the suturing needle usually requires coordination between the two manipulators to allow correct grasping. As the CHENA facilitates this, 
we simply ensure that the \textit{Hand-off} occurs in proximity of the next entry point, facilitating the next \textit{Positioning} while allowing the PSM1 manipulator to freely approach the PSM2.
\vspace{-2pt}
\begin{gather}\label{eq. pull}
\tau= ( \tau^{x_{j+1}},\tau_h^y, \tau^z ) \,,\quad
\lambda = ( \lambda^x, 0, \lambda^z )
\end{gather}

\subsubsection{Other}

When the user's kinematic inputs to the PSMs do not match any of the first four surgemes, no constraint is applied to the manipulator, which corresponds to $\lambda=0$ for all Cartesian coordinates.
\vspace{-2pt}
\begin{gather}\label{eq. positioning}
\tau= ( \tau_h^x,\tau_h^y, \tau_h^z ) \,,\quad
\lambda = ( 0, 0, 0 )
\end{gather}

Notice that it would be possible to set hard constraints to impose forbidden movements to the user for each control paradigm, but interaction forces with the human may disrupt the control. Instead, the C-IAC allows for the creation of softer and more dynamic constraints by predicting the intent and considering the system's confidence throughout the task.

\begin{figure}[h] 
\centering 
\includegraphics[height=5.3cm, trim={5.1cm 2.1cm 5.1cm 1.9cm },clip]{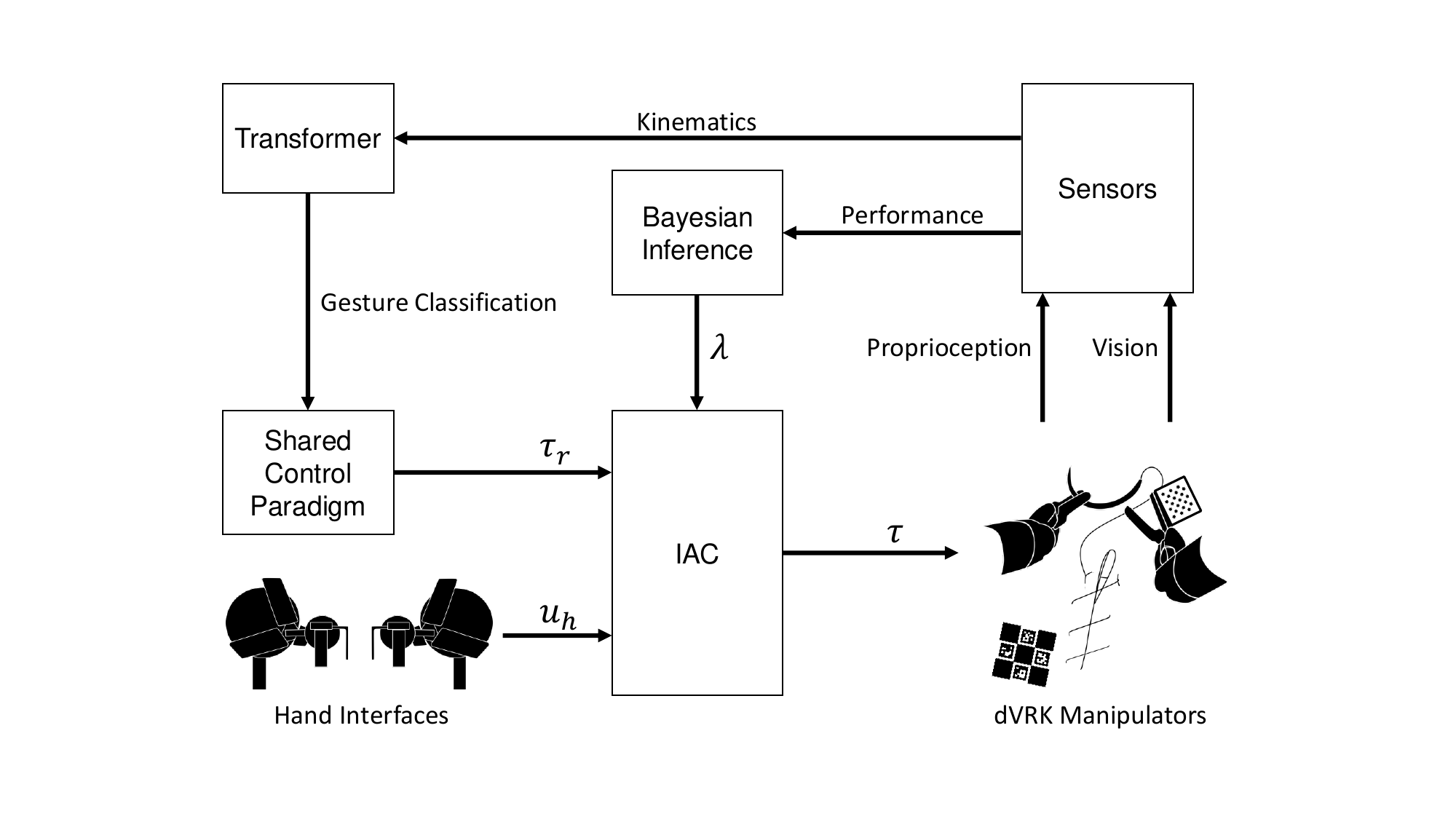}
\caption{Bimanual suturing control scheme using C-IAC and Transformer for gesture classification. Sensory information from markers and robot kinematics undergoes Bayesian inference to compute the confidence parameter $\lambda$ for the IAC. 
The Transformer utilizes the kinematics to classify the user’s gesture and define the target $\tau_r$. In IAC, the user input $u_h$ is used to predict the human's target $\tau_h$, which, together with $\tau_r$, determines the robot manipulator's target.} 
\label{fig: logic_chart_CIAC}
\end{figure}

The complete system framework is shown in Fig.\,\ref{fig: logic_chart_CIAC} .
\vspace{-5pt}

\section{Experiments}
The experiments were approved by the College Research Ethics Committee of Imperial College London (21IC7042). Participants were briefed on the experiments' purpose and protocol, and signed a consent form before starting.
The first set of experiments was carried out to validate our Transformer-based model, the STITCHES dataset, and needle pose estimation accuracy.
Subsequently, we conducted two user studies to compare traditional teleoperation against C-IAC in two tasks: (i) target reaching, to systematically analyze the controller performance, and (ii) four surgical throws, to represent a key surgical function.
For consistency, the PSM1 was equipped with a CHENA in all experiments. This setup implies that while C-IAC controls both PSMs, only the one with the keydot uses confidence-based properties, since it is tracked both kinematically and visually.

\subsection{Target Reaching User Study}
To perform a detailed comparison between C-IAC and traditional teleoperation without the complexities of a complete suturing task, we design a target reaching experiment consisting of \textit{Positioning} and \textit{Push}. 
We recruited 8 participants (1 female and 7 males, age 23-30, all right-handed) without dVRK experience. 
After 10 minutes of familiarisation, the subjects controlled the PSM1 to reach four entry points on the phantom, starting from a fixed pose. In each trial, the experimenter triggered a button to indicate successful completion of reaching and needle insertion. 
The starting pose remained consistent, with entry points positioned on a straight line at \{15, 30, 45, 60\}mm from the starting pose, thus challenging the user's depth perception.

We compared traditional teleoperation versus C-IAC, each tested with four subjects.
To ensure consistency across participants, $\lambda$ in C-IAC was gradually increased from 0 to 0.8 using a linear function instead of Bayesian Inference. This ensures uniform assistance across trials.
We then use Equation \ref{eq. push} as the control paradigm during all C-IAC trials.
This and $\lambda=0.8$ were selected pragmatically to ensure the human retains a degree of control throughout the experiment.

\subsection{Suturing Task User Study}

This study compared the proposed C-IAC against traditional teleoperation in a full suturing task with four entry points that require four suturing throws.
The experiment aimed to assess the performance of seven novice users (age 21-29, no medical background), as well as one intermediate (age 23, fifth-year medical student) and one expert (age 41, neurosurgeon with practice in the medical field). None had experience with surgical robots.
Before the experiments, each subject trained on the system for eight sessions of 20 minutes each.

A clinical criterion for successful suturing is maintaining  the needle plane perpendicularity to the wound and skin \cite{guni2018development, pedram2020autonomous}. Since the CHENA keeps the needle perpendicular to the manipulator, we tracked its orientation via the keydot marker to measure the needle perpendicularity during the task.
Additionally, we measured the Transformer's average accuracy, task completion time, and participants' perceived task load through a NASA-TLX questionnaire.




\section{Results}

\begin{figure*}[ht!] 
\centering 
\includegraphics[height=8.8cm, trim={0cm 0.5cm 0cm 1.3cm},clip]{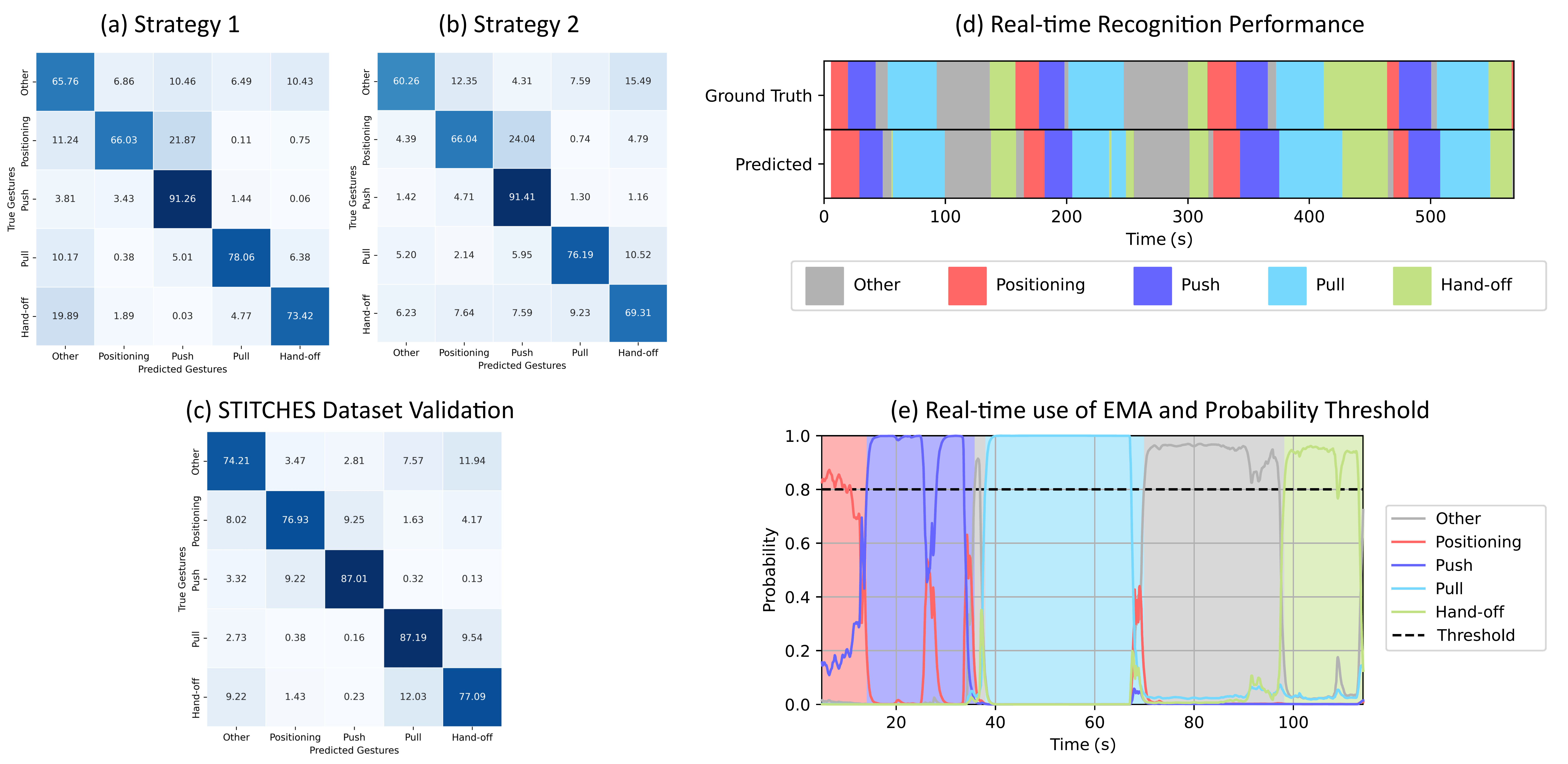}
\caption{Transformer model classification on JIGSAWS and STITCHES datasets. (a,b) Transformer gesture classification using labeling strategies 1 and 2. (c) 5-fold cross validation average accuracy of STITCHES. The overall accuracy is 81.19\% using the proposed labeling. (d) Ground truth and predicted gesture comparison with STITCHES. (e) Real-time suturing throw with EMA and probability threshold.} 
\label{fig: validation_and_threshold}
\end{figure*} 

\subsection{Transformer for Two Classification Strategies}

The Transformer model is applied to two strategies in JIGSAWS (Section \ref{Sec.: Transformer}) to demonstrate the feasibility of reducing the suturing task into four fundamental motions and a general one. Fig.\,\ref{fig: validation_and_threshold} (a-b) compares their confusion matrices. 
The model achieves an overall accuracy of 77.67\% and 76.72\% for strategy 1 and 2, respectively.
Both show consistent trends, with some misclassifications in \textit{Positioning} and \textit{Push}, which are both predominantly right-hand gestures. \textit{Other} has the lowest accuracy in both strategies, while \textit{Push} was the highest.
The comparable results confirm that strategy 2 effectively generalizes the task, making it the basis for subsequent experiments.

\subsection{STITCHES Dataset Validation}
Training and validation of our Transformer model were conducted on STITCHES, which was then used in the user studies. 
Using post-processing techniques, i.e. EMA and probability threshold, the model achieved an accuracy of 81.19\%. Fig.\,\ref{fig: validation_and_threshold} (c-d) shows the confusion matrix obtained from validation and a comparison between the actual gestures and the predicted in a sample. Additionally, Fig.\,\ref{fig: validation_and_threshold} (e) provides a graphical representation of the post-processing applied in real-time during a suturing throw.


\subsection{Needle Pose Estimation Error}

With the CHENA, the needle plane can be approximately considered in a fixed orientation perpendicular to the gripper, similarly to \cite{sen2016automating, pedram2020autonomous}. To compute the needle pose error, we compared the CHENA's pose estimation from visual tracking of the keydot with the robot's kinematic data. Table \ref{table: pose estimation error} presents the mean and standard deviation of position and orientation errors across 80 random robot positions, showing errors in the order of 1\,mm and 2$^\circ$.
\vspace{-2pt}
\begin{table}[!h]
\centering

\caption{\label{table: pose estimation error} 
Needle Pose Estimation Error Results}
\begin{tabular}{ccccccc}
\hline
\multirow{2}{*}[1.5ex]{\text{  } } & \multicolumn{3}{c}{\textbf{Position (mm)}} & \multicolumn{3}{c}{\textbf{Orientation ($\circ$)}} \\
\hline \textbf{DoF}  &
\textbf{x} &
\textbf{y} &
\textbf{z} & 
\textbf{Roll} &
\textbf{Pitch} &
\textbf{Yaw}  \\
\hline Mean & 0.652 & 0.933 & 0.843 & 2.510 & 2.299 & 1.238 \\
Std. & 0.997 & 0.873 & 0.676 & 2.727 & 1.828 & 1.375 \\
\hline
\end{tabular}
\end{table}



\subsection{Target Reaching User Study}

During this study, we measured the time required for the reaching movement to perform accurate insertions.
As the data was not normally distributed, a Wilcoxon rank-sum test was conducted, revealing a difference between the controllers ($p=0.004$).
We see in Table \ref{table: reaching time} that for all entry points, C-IAC reduced the average time required for target reaching and precise insertion relative to traditional teleoperation, with a total average and standard deviation of ($\sigma, \mu$) = ($36.1, 15.4$)\,s, and ($61.9, 39.6$)\,s for traditional control.
\vspace{-2pt}
\begin{table}[!h]
\centering

\caption{\label{table: reaching time} 
Duration of Target Reaching (in seconds)}
\resizebox{\columnwidth}{!}{ 
\begin{tabular}{cccccc}
\hline \textbf{Controller}  & \textbf{Entry point 1} & \textbf{Entry point 2} & \textbf{Entry point 3} & \textbf{Entry point 4} & \textbf{Total} \\
\hline Traditional & $61\pm6.8$ & $61\pm33.3$ & $41\pm10.6$ & $84.8\pm72.8$ & $61.9\pm39.6$ \\
\hline C-IAC & $25\pm5.4$ & $45\pm24.4$ & $39.5\pm13.5$ & $35\pm9.8$ & $36.1\pm15.4$  \\
\hline
\end{tabular}
}
\end{table}

\begin{table*}[th!]
\centering
\caption{\label{table: suturing_study} 
User Study Suturing Task Performance based on Participant Suturing Experience
}
\begin{tabular}{|c|cc|cc|cc|}

\hline

\multirow{2}{*}[1.5ex]{\text{  } } &  \multicolumn{2}{c|}{\textbf{Novices}} & \multicolumn{2}{c|}{\textbf{Intermediate}} & \multicolumn{2}{c|}{\textbf{Expert}} \\
\cline{2-7} \textbf{\text{  }}  &
\textbf{Traditional} &
\textbf{C-IAC} &
\textbf{Traditional} & 
\textbf{C-IAC} & \textbf{Traditional} & \textbf{C-IAC} \\

\hline Avg. $\perp$ Error in Push ($\circ$) \hspace{\fill}\hspace{\fill} & 28.83$\pm$10.11 & 7.27$\pm$5.26 & 21.98 & 4.25 & 34.47 & 2.19\\
\hline Real-time Classification Accuracy (\%)\hspace{\fill} & --- & 65.50$\pm$15.25 & --- & 67.37 & --- & 67.17\\
\hline Avg. Positioning Time (s) \hspace{\fill} & 26.31$\pm$18.99 & 19.94$\pm$9.23 & 24.82 & 22.87 & 36.95 & 32.93\\
\hline Avg. Push Time (s) \hspace{\fill} & 43.07$\pm$18.97 & 38.05$\pm$21.95 & 42.00 & 47.83 & 47.33 & 40.07\\
\hline Avg. Pull Time (s) \hspace{\fill} & 76.91$\pm$40.68 & 48.26$\pm$17.35 & 65.65 & 50.97 & 87.10 & 66.12\\
\hline Avg. Hand-off Time (s) \hspace{\fill} & 54.09$\pm$19.21 & 35.29$\pm$14.17 & 36.86 & 27.68 & 74.84 & 49.66\\
\hline Avg. Throw Time (s) \hspace{\fill} & 222.70$\pm$73.80 & 155.94$\pm$40.41 & 214.66 & 181.15 & 290.27 & 218.46\\
\hline Total Suturing Time (s) \hspace{\fill} & 892.18$\pm$204.53 & 652.55$\pm$158.71 & 859.01 & 724.95 & 1161.51 & 873.87\\
\hline
\end{tabular}
\end{table*}

\subsection{Suturing Task User Study}

\begin{figure*}[th!] 
\centering 
\includegraphics[height=5.5cm, trim={16cm 5.6cm 15cm 4.3cm},clip]
{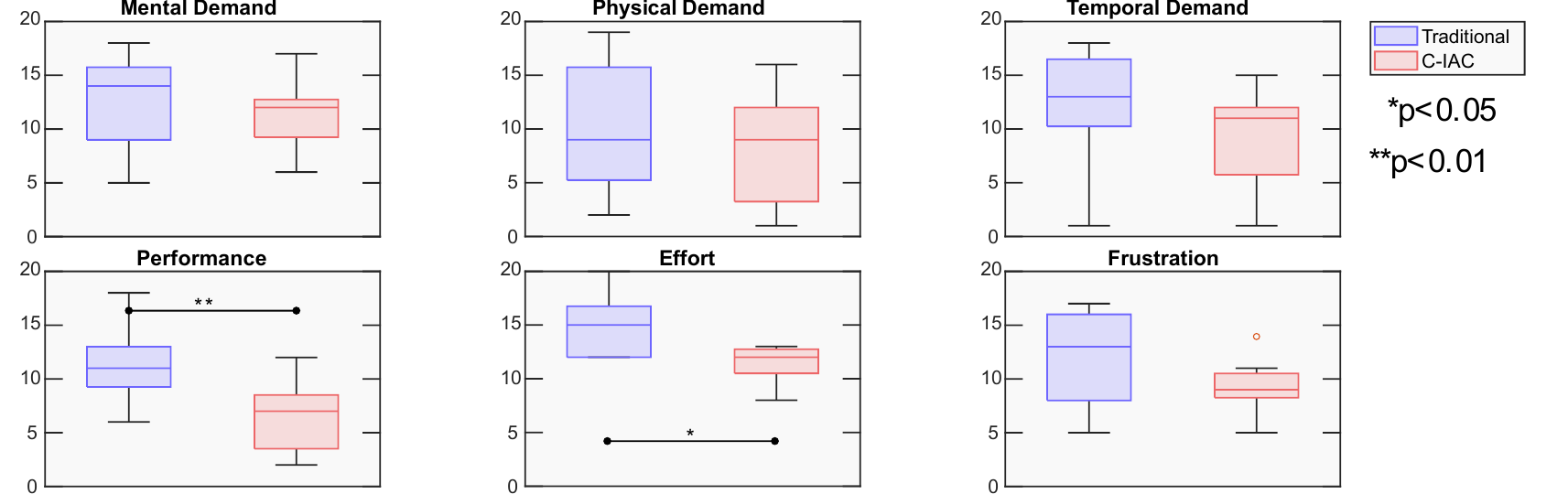}
\caption{Subjective assessment of C-IAC vs traditional teleoperation with NASA-TLX with the seven novice
participants after four throws suturing user study. Lower values indicate better performance.} 
\label{fig: NASA_TLX_paper_spider_plots}
\end{figure*}

During the study with users of different suturing skill levels, participants performed four consecutive throws. 
The results in Table \ref{table: suturing_study} highlight C-IAC's smaller perpendicular error and reduced task duration across all surgemes relative to traditional teleoperation. 
As the seven novices' data is not normally distributed, Wilcoxon Signed Rank test was used, which revealed a significant reduction in perpendicular error during \textit{Push} with C-IAC ($p<0.001$). Furthermore, C-IAC yields faster performance than traditional teleoperation ($p<0.001$) in singular throw time, in particular during \textit{Pull} ($p<0.001$) and \textit{Hand-off} ($p<0.001$).
Interestingly, the intermediate and expert surgeon exhibit similar trends.
The NASA-TLX responses from the novices (Fig.\,\ref{fig: NASA_TLX_paper_spider_plots}) indicate a lower perceived workload with C-IAC. After confirming data normality, we performed t-test to compare the scores, summarized in Fig.\,\ref{fig: NASA_TLX_paper_spider_plots}. 
The analysis found that participants felt they performed better with C-IAC ($p<0.01$) and required less effort ($p<0.05$).
Similar trends were observed for temporal demand ($p=0.079$) and frustration ($p=0.083$).

\section{Discussion}

The classification accuracy results confirmed the feasibility of using four fundamental surgemes and a general one. While \textit{Push} had the highest overall accuracy, \textit{Other} had the lowest, likely due to the aggregation of diverse gestures.
To train the Transformer, we created STITCHES and implemented strategy 2, which yielded a similar accuracy trend as with JIGSAWS. 

In creating STITCHES, we observed bimanual hand movements to pull enough thread for subsequent throws for the first 1-2 throws, as seen in Figure \ref{fig: validation_and_threshold} (d-e), where \textit{Other} follows \textit{Pull}. While this movement is not fundamental to a single stitch, it is essential for consecutive throws. Our online gesture recognition more accurately reflects the real task than \cite{pasini2023grace}, where this property was missed due to using only two throws or short thread.
Needle pose estimation is consistent with the literature \cite{sen2016automating, pedram2020autonomous} and sufficiently accurate to carry out the experiments.

In the first user study, C-IAC proved to be faster than traditional teleoperation, leading to a smaller standard deviation. 
We conducted a second user study involving four consecutive throws with participants of different skill levels, without prior experience in robotic suturing.
C-IAC enabled faster task completion than traditional teleoperation, both in average throw and total task time. For needle insertion, C-IAC significantly reduced perpendicularity error while maintaining similar \textit{Push} times. 
Additionally, it improved performance in \textit{Pull}, \textit{Hand-off}, and throw time, demonstrating seamless integration of the autonomous component into the workflow.

In our experiments, the model showed lower accuracy than during dataset validation, as STITCHES was recorded from a different user. Additionally, the intermediate and expert, having already developed their own suturing styles, were more affected than novices.
Furthermore, the participants required more suturing time than the demonstrations in STITCHES, impacting classification accuracy.
Notably, the expert took the longest, perhaps due to a bias towards non-robotic laparoscopic suturing influencing the learning curve \cite{kaul2006learning}.

Fig.\,\ref{fig: NASA_TLX_paper_spider_plots} reports NASA-TLX scores given by the novices, indicating that C-IAC imposes less workload than traditional control, with significant improvements in performance and effort. 
The higher suturing time compared to previous works \cite{gao2014jhu, sen2016automating} reflects the challenging surgical setting designed to encourage human-robot interaction. Latency, 2D display, different hand interfaces, and inexperience with the surgical setup are the main contributors to this increased time.

C-IAC relies on a trained gesture recognition model, which itself requires a dataset. To avoid circular dependence, C-IAC's Transformer model was trained on traditional teleoperation movements.
This solution demonstrates the C-IAC ability to handle lower accuracy models by adjusting the confidence $\lambda$ parameter.
In our tests, setting $\lambda$ below 1 proved preferable due to human and robot sensor limitations.
This suggests that enhanced sensory input, such as additional cameras, force sensors, or advanced path planning, could enable higher confidence (i.e. $\lambda=1$), potentially improving the performance.

\section{Conclusion}

At the extremes of the control spectrum, traditional teleoperation faces delay and sensory limitations, while full automation is limited by sensory inaccuracies and lack of broad task knowledge. These challenges are addressed here by developing a unified human-robot system with high and low level intent detection to improve suturing performance while relieving cognitive workload.
To our knowledge, this represents the first implementation of real-time intent detection for the interaction control in bimanual robotic suturing, considering confidence across multiple sensory modalities.
The proposed teleoperation system allows the operator to perform more efficient gestures, with reduced cognitive strain, which can be easily extended to other fields requiring remote teleoperation.
In the future, a larger user sample size should be acquired to analyze the C-IAC and search for optimal probability thresholds and $\lambda$ limits.

\section*{Acknowledgments}
This work was conducted at Imperial College London's Hamlyn Centre for Robotic Surgery. Special thanks to Kiran Bhattacharyya and Anton Deguet for their support.

\bibliographystyle{IEEEtran}
\bibliography{IEEEabrv,bibliography}

\vfill

\end{document}